\documentclass{bmvc2k}

\title{Robust Ensemble Model Training via Random Layer Sampling Against Adversarial Attack}

\addauthor{Hakmin Lee\textsuperscript{*}}{zpqlam12@kaist.ac.kr}{1}
\addauthor{Hong Joo Lee\textsuperscript{*}}{dlghdwn008@kaist.ac.kr}{1}
\addauthor{Seong Tae Kim}{seongtae.kim@tum.de}{2}
\addauthor{Yong Man Ro}{ymro@kaist.ac.kr}{1}

\addinstitution{
 Image and Video Systems Lab, \\ School of Electrical Engineering, \\ KAIST, South Korea
}
\addinstitution{
 Computer Aided Medical Proc-\\edures, Technical University of\\ Munich, Germany
}

\runninghead{Lee et al.}{Robust Ensemble Model Training}

\usepackage{amsmath,amssymb}
\usepackage{multirow}
\usepackage{multicol}

\DeclareMathOperator{\E}{\mathbb{E}}
\begin{document}

\maketitle

\begin{abstract}
Deep neural networks have achieved substantial achievements in several computer vision areas, but have vulnerabilities that are often fooled by adversarial examples that are not recognized by humans. This is an important issue for security or medical applications. In this paper, we propose an ensemble model training framework with random layer sampling to improve the robustness of deep neural networks. In the proposed training framework, we generate various sampled model through the random layer sampling and update the weight of the sampled model. After the ensemble models are trained, it can hide the gradient efficiently and avoid the gradient-based attack by the random layer sampling method. To evaluate our proposed method, comprehensive and comparative experiments have been conducted on three datasets. Experimental results show that the proposed method improves the adversarial robustness. 
\end{abstract}

\section{Introduction}
\label{sec:intro}
	Recently, deep learning models have shown exceptionally good performance in various computer vision tasks such as image classification \cite{simonyan2014very, he2016deep}, object detection \cite{ren2015faster, ren2016object}, semantic segmentation \cite{long2015fully, chen2017deeplab}. However, several studies have revealed that deep learning methods are vulnerable in case images are intervened by small perturbations that are not even perceptible for human-beings. These perturbed images are called adversarial examples, and the procedure to create such adversarial examples is called adversarial attack. 

The adversarial attack algorithms could be categorized into two approaches. The one is white-box attack and the other one is black-box attack. In the white-box attacks, the attacker could access to the model's parameters, while in the black-box attacks, the attacker could not access to the model's parameters. In this paper, we focus on handling white box attacks. In the white-box attacks, interruptions of visually imperceptible perturbations to original images could lead to erroneous results. The perturbations of these approaches are generated based on gradients. The gradient-based adversarial attack methods could simply and effectively cause malfunctions of target networks. Hence, the presence of adversarial examples has brought up challenges of great importance for security-critical computer vision applications.

To enhance the deep learning model against adversarial examples, several methods have been proposed. Previous adversarial defense methods have been proposed: Randomization \cite{ pang2019improving} to make the attack ineffective, denoising \cite{das2018shield}, ensemble training \cite{pang2019improving}, and training the model with adversarial examples \cite{madry2018towards, athalye2018obfuscated}. It has been demonstrated that training with adversarial examples is the most effective way to improve the model robustness \cite{ athalye2018obfuscated}. However, it takes large training time and impractical on large-scale datasets \cite{wang2020improving}. 

To increase the robustness of the deep network, in recent studies, randomization and obfuscation are reported as practically effective strategies to improve the adversarial robustness. The example of randomization is noise addition at different levels of the system \cite{you2019adversarial}, randomized lossy compression \cite{das2018shield}, random projections \cite{vinh2016training}, and random feature sampling \cite{chen2019secure}. The key idea of these approaches is to hide the gradient of the network by randomization. The adversarial attack is blunted in white box situations, where the defender can effectively perturb the weight through randomness and make a difference when the attacker makes adversarial examples based on a specific weight. In other words, increasing the diversity of the networks plays an important role in defense.

One of the simple ways to increase the model diversity and improve the robustness is to generate ensemble models. It has been empirically observed by Smith et al. \cite{lakshminarayanan2017simple} that ensemble networks which are trained with different random initialization can be robust to adversarial examples. Pang et al. \cite{pang2019improving} proposed an ensemble model training method to promote the diversity among the predictions. They proposed the adaptive diversity promoting regularizer to make individual network predict orthogonally. Most of these ensemble-based approaches showed prominent in terms of adversarial robustness. However, these ensemble models require a large number of parameters to improve the adversarial robustness.

In this paper, we focus on tackling the problem of network parameter increase in the ensemble models when improving the adversarial robustness. To this end, we propose a novel ensemble model training framework with random layer sampling and group optimization strategy. In the proposed ensemble models training framework, the ensemble model set is defined with \textit{M} sub-models. Each sub-model has same structure with \textit{L} layers and they have different weights. From the model set, we sample the layers categorically through the proposed random layer sampling method. Then, we generate sampled models which have the same structure with sub-model. Each sampled model predicts sample outputs. Through the proposed random layer sampling method, it is possible to increase possible recombination cases exponentially by simply adding linear parameters. Also we train the ensemble models with group optimization to promote the ensemble diversity. Through the group optimization, we could predict diverse sample outputs that are robust to adversarial examples.
In summary, there are two advantages of the proposed method. Firstly, we can generate various ensemble models combination only with a few number of sub-models. Secondly, since our proposed layer sampling method generates a sampled model randomly, we can effectively take a gradient ambiguity and avoid reproducible for the adversarial attack. Then, the sample outputs guarantee the diversity. Therefore, we could predict robust prediction against adversarial examples. The contributions of our paper can be summarized below:

\begin{itemize}
	\item We propose a novel ensemble model training framework with random layer sampling. Through the proposed framework, we can effectively construct $M^L$ ensemble models with \textit{M} sub-models which has \textit{L} layers. Compared to conventional ensemble methods, our method can generate a large number of predictions with a few number of sub-models. Then, the diversity of the predictions is increased with the random layer sampling training framework and group optimization strategy.
	
	\item Our method effectively hides the full gradient of network and improves the adversarial robustness. It is mainly due to the reason that our method predicts various sample outputs from different model combinations. Experimental results show that our method effectively improves the adversarial robustness compared with other ensemble-based defense methods. 
\end{itemize}

\section{Related Work}
\label{sec:related}
\subsection{Adversarial Attack Method}
The deep neural networks have been shown to be highly vulnerable to adversarial examples. It was first discovered by Szegedy et al. \cite{szegedy2013intriguing}. Then, Goodfellow et al. \cite{goodfellow2014explaining} proposed Fast Gradient Sign Method (FGSM). It is a fast and single-step adversarial attack version of \cite{szegedy2013intriguing}. It performs a single step update on the original sample $x$ along the direction of the gradient of a loss function. After that, Moosavi et al. designed the DeepFool attack \cite{moosavi2016deepfool} starting from the assumption that models are fully linear. Under this assumption, there is a polyhedron that can separate individual classes. Recently, more powerful and effective attacks including C\&W \cite{carlini2017towards}, PGD \cite{madry2018towards}, EAD \cite{chen2018ead} are proposed to fool the networks. As stronger attack methods are reported, the need for developing better defense methods is increased.
\subsection{Adversarial Defense Method}
To improve the model's robustness against adversarial examples, several methods have been reported. There are many approaches including distillation-based approaches \cite{papernot2016distillation}, adversarial training approaches \cite{madry2018towards, athalye2018obfuscated}, and ensemble training approaches \cite{lakshminarayanan2017simple, pang2019improving, liu2018towards, dhillon2018stochastic}. In this section, we mainly describe the ensemble training approaches.

Pang et al. \cite{pang2019improving} improves the adversarial robustness by promoting ensemble diversity with Adaptive Diversity Promoting (ADP) regularization. They trained the ensemble model through the ADP regularization method. Through the ensemble diversity, the non-maximal predictions of each model are mutually orthogonal, then predicts robust output. Smith et al. \cite{smith2018understanding} proposed ensemble models training method with predictive uncertainty estimation and detecting adversarial example. By quantifying the predictive uncertainty, they could optimize the ensemble models that are robust to adversarial examples. It could also detect adversarial examples with predictive uncertainty. Although these ensemble model training methods could improve the adversarial robustness, to improve the adversarial robustness, a large number of parameters are required. Another way to construct ensemble models is to add the noise in the layer or dropout the weight. The noise addition or dropout method can be interpreted as ensemble model combination \cite{lakshminarayanan2017simple} where the predictions are averaged over an ensemble of neurons. Liu et al. \cite{liu2018towards} reported a random self-ensemble (RSE) training for improving the robustness of deep neural models. They add a noise layer before each convolution layer in both training and prediction phases. They showed that the algorithm is equivalent to ensemble a huge amount of noisy models together, and ensure that the ensemble model can generalize well. They further prove the fact that the proposed method is equivalent to adding a Lipchitz regularization and thus can improve the robustness of neural models. Dhillon et al. \cite{dhillon2018stochastic} proposed stochastic activation pruning for robust adversarial defense. During the forward pass, they prune a subset of the activations in each layer. Then, they scale up the remaining activations to normalize the dynamic range of the inputs to the subsequent layer.
\begin{figure*}[t]
	\centering
	\includegraphics[width=0.9\textwidth]{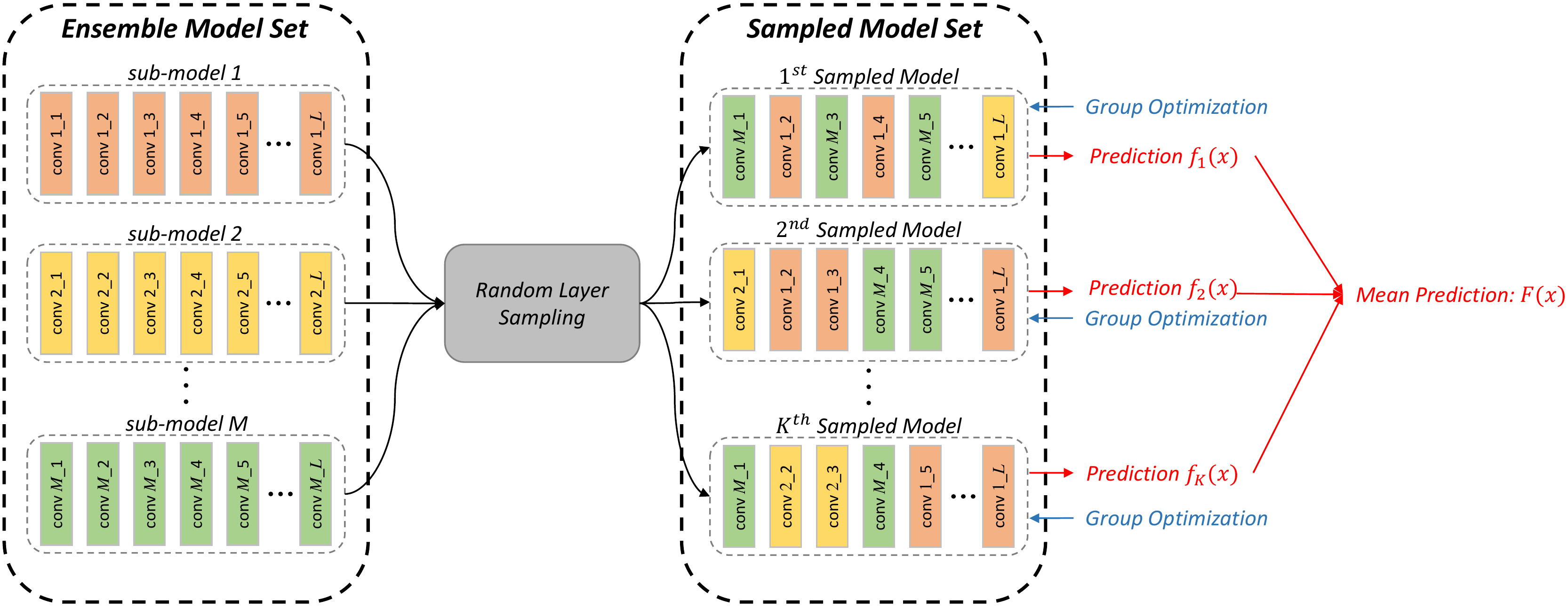}

	\caption{Overall procedure of the proposed ensemble model training framework with random layer sampling method. The ensemble model set consists of \textit{M} sub-models. Each sub-model has \textit{L} layers with different weights. From the ensemble model set, we sample each layer and construct sampled model. Then, we construct \textit{K} different sampled models and update each sampled model parameter at once.}
	\vspace{-0.5cm}
	\label{fig1}
\end{figure*} 

\section{Proposed Method}
\label{sec:Method}
In this section, we describe our proposed ensemble model training framework. Figure 1 shows an overview of the proposed ensemble model training framework with random layer sampling and group optimization. As shown in the figure, we construct ensemble model set with \textit{M} duplicate sub-models. The sub-model consists of \textit{L} layers, and weight of each layer is differently initialized over sub-models. To train the ensemble model set, we sample the layer from the ensemble model set by the proposed random layer sampling method. Then, we construct \textit{K} sampled models. 
The sampled model predicts sample output $f_{k}(x)$. From the sampled models, we optimize the sampled models at once with sample mean prediction to increase the diversity between sampled models. In the testing phase, we construct \textit{N} different ensemble models by the proposed random layer sampling and decide a final decision by averaging predictions of \textit{N} sampled models. The details of the proposed random layer sampling method and training framework will be described in the following subsections.
\subsection{Random Layer Sampling}
In this section, we describe how to sample the layer and generate a sampled model. As shown in Figure 1, we design an ensemble model set with \textit{M} duplicate sub-models which consists of \textit{L} layers. Each sub-model has the same structure while having different weights. From the ensemble model set, we sample the layers by our proposed Random Layer Sampling (RLS) method. In the proposed RLS method, we sample layers with $r_{l}(m)$ where \textit{l} denotes the layer index, \textit{m} denotes the sub-model index, and $r_{l}(m)$ denotes the categorical random variable that indicates whether each layer is selected or not. In order to ensure that each layer is selected from one sub-model, we constrain the $r_{l}(m)$ as follows

\begin{equation}\label{eq1}
\begin{aligned}
\sum_{m=1}^{M}r_{l}(m)=1,
\end{aligned} 
\end{equation}

\begin{equation}\label{eq2}
\begin{aligned}
r_{l}\sim categorical(x_{1},x_{2},x_{3}, \dots, x_{M}; \mu_{1},\mu_{2},\mu_{3}, \dots, \mu_{M}). 
\end{aligned} 
\end{equation}

Therefore, the feed-forward operation can be described as

\begin{equation}\label{eq3}
\begin{aligned}
w_{l}=r_l\otimes \mathbf{W}_l,
\end{aligned} 
\end{equation}

\begin{equation}\label{eq4}
\begin{aligned}
z^{(l+1)}=w_l*y^{(l)},
\end{aligned} 
\end{equation}

\begin{equation}\label{eq5}
\begin{aligned}
y^{(l+1)}=\sigma(z^{(l+1)}),
\end{aligned} 
\end{equation}
where $\mu_{M}$ denotes probability of each category separately specified, $\mathbf{W}_{l}$ denotes a set of weights in $l^{th}$ convolution layer, $w_l$ denotes a sampled weight from categorical random variables, $\otimes$ denotes element-wise multiplication, * denotes a convolution operator, $y^{l}$ denotes a feature vector of $l^{th}$ layer, $z^{(l+1)}$ denotes the output vector of $l^{th}$ layer, and $\sigma$ is an activation function. Eq. 3 denotes that we sample only one weight of the layer in the \textit{M} sub-models. By stacking these randomly sampled layers, we can construct sampled models which have the same structure but have different weights. Through the RLS, we can generate various ensemble models that have different weight effectively. Theoretically, we can generate $M^L$ different ensemble models only with the \textit{M} sub-models.
\subsection{Training for Adversarial Robustness with Ensemble Diversity}
It is widely known that ensemble of several individual models could improve the adversarial robustness \cite{lakshminarayanan2017simple, pang2019improving, liu2018towards, dhillon2018stochastic}. For the adversarial robustness, the diversity among individual sub-models should be sufficiently guaranteed.
To guarantee the diversity, we trained the ensemble models by group optimization strategy. As shown in Figure 1, let $f_{k}(x)=p(y\mid x,\hat{w}^{k})$ be a prediction of the sampled model where $\hat{w}^{k}$ denotes the sampled weights, and $F(x)=\frac{1}{K}\sum{f_k(x)}$ be a mean prediction of the sampled models. Following the description of \cite{krogh1995neural, zhang2019confidence}, the ensemble model diversity can be defined as
\begin{equation}\label{eq6}
\begin{aligned}
\alpha (f_k \mid x)=(f_k(x)-F(x))^2.
\end{aligned} 
\end{equation}
Therefore, the diversity of the ensemble model can be defined as the difference between individual sub-model prediction and mean prediction of sampled model. If we set the difference between ground-truth and sample output as mean square error, it can be represented as $MSE(f_k \mid x) = (y-f_k(x))^2$. The mean square error can be decomposed into
\begin{equation}\label{eq7}
\begin{aligned}
\E [MSE(F \mid x)] = \E [\overline{MSE}(f \mid x)] - \E [\bar{\alpha}(f \mid x)],
\end{aligned} 
\end{equation}

where

\begin{equation}
\begin{aligned}
\overline{MSE}(f \mid x) = \frac{1}{K}\sum_{k=1}^{K} MSE(f_k \mid x), \;and\; \bar{\alpha}(f \mid x) = \frac{1}{K}\sum_{k=1}^{K} \alpha(f_k \mid x).
\end{aligned} 
\end{equation}
%
%
%
To minimize sample mean prediction, we group \textit{K} different sampled models with the proposed random layer sampling method. By minimizing $\E [MSE(F \mid x)]$, the set of individual sampled model is optimized and the diversity is guaranteed. 



\subsection{Adversarial Defense Scenario}
\begin{figure*}[t]
	\centering
	\includegraphics[width=0.9\textwidth]{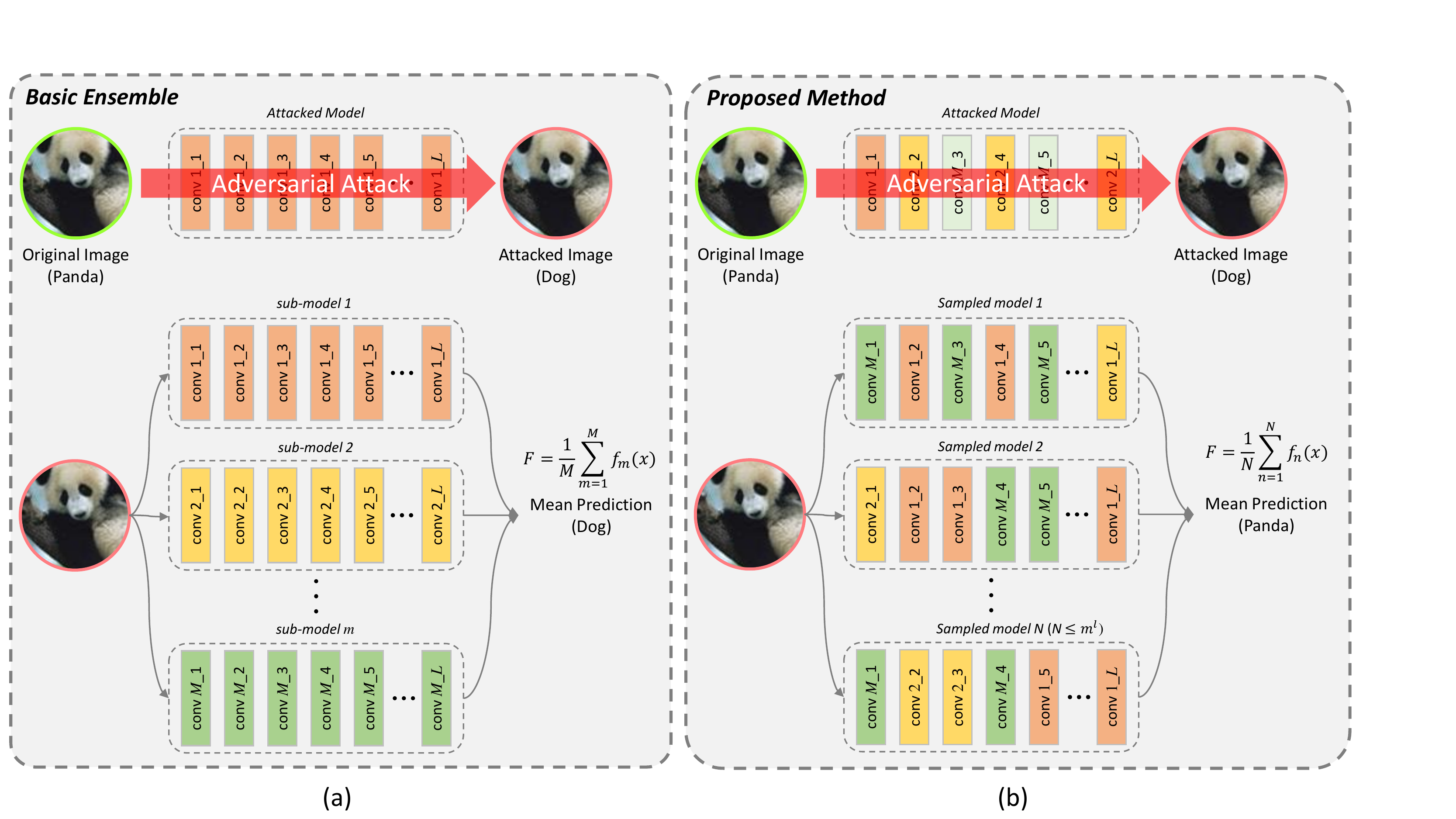}

	\caption{Comparison of defense scenarios between (a) basic ensemble method and (b) our method.}
	\vspace{-0.5cm}
	\label{fig2}
\end{figure*} 
In this section, we describe the adversarial defense scenario. Figure 2 shows the comparison of defense scenario between basic ensemble method and our method. As shown in Figure 2 (a), in the basic ensemble model, if the sub-model is attacked, the prediction is corrupted by the adversarial attack. Note that the prediction score is calculated by averaging \textit{M} predictions. To reduce the effect of adversarial attack, it is required to increase the number of sub-models but it is limited in real-world applications.

Compared to the basic ensemble approach, our method could generate $M^L$ different ensemble models effectively only with \textit{M} sub-models. As shown in Figure 2 (b), although a sampled model is attacked, at the test time, our method randomly samples \textit{N} different models with the RLS method. It is very low probability to sample exactly same sub-model which is attacked $(1/M^L)$. Since the weight of the attacked model are different from the ones of the sampled models, the gradients of the attacked model is different from sampled models. Therefore, the attack algorithm does not work properly in the sampled models. As a result, the robustness is improved in the proposed method.

\section{Experiments}
\subsection{Datasets and Implementation Details}
To verify the effectiveness of the proposed method, we use three widely studied datasets-MNIST \cite{lecun1998gradient}, CIFAR10 \cite{krizhevsky2009learning}, and SVHN \cite{netzer2011reading} dataset. MNIST dataset is a collection of handwritten digits in classes 0 to 9. It consists of 60,000 training images and 10,000 test images. SVHN dataset is similar to MNIST dataset but colorful street view house numbers. It has 10 classes and consists of 63,257 training images and 26,032 test images. CIFAR10 dataset consist of 60,000 images with 10 classes. It consists of 50,000 training images and 10,000 test images. Each class has 6,000 images. The pixel value of images is scaled to be in an interval [-1, 1].
We evaluate our method with two publicly available networks (ResNet-50 \cite{he2016deep} and VGG-16 \cite{simonyan2014very}). To verify that the proposed method is compact while robust to adversarial attack, we only use two sub-models and 3 sampled models for training (\textit{M}=2, \textit{K}=3). All networks are trained from the scratch by using SGD optimizer. The initial learning rate is set as 0.01 and divided by 10 every 50 epochs. We use a weight decay of $5 \times 10^{-4}$ and a momentum of 0.9. We run the training process for 50 epochs on MNIST, 150 epochs on CIFAR10 and SVHN.
\subsection{Performance Comparison under Adversarial Attacks}
\begin{table}[t]
	\centering
	\caption{Classification accuracy (\%) on adversarial examples. Models consist of Resnet-50.}
	\scalebox{0.8}{
	\begin{tabular}{c|c|c|c|c|c|c|c|c|c}
		\hline
	& \multicolumn{3}{c|}{\textbf{MNIST}} & \multicolumn{3}{c|}{\textbf{CIFAR-10}} & \multicolumn{3}{c}{\textbf{SVHN}}  \\
		Attack Method               & Para.    & Baseline & Ours           & Para.     & Baseline  & Ours            & Para.    & Baseline & Ours           \\ \hline\hline
		No Attack & - & 99.5 & \textbf{99.6} & - & 95.1 & \textbf{96.7} & - & 96.2 & \textbf{97.6} \\ \hline
		\multirow{2}{*}{FGSM} & $\epsilon$=0.1  & 89.5     & \textbf{98.6} & $\epsilon$=0.01  & 45.4      & \textbf{65.1}  & $\epsilon$=0.05 & 40.4     & \textbf{56.2} \\
		& $\epsilon$=0.2  & 44.1     & \textbf{70.4} & $\epsilon$=0.02  & 35.6      & \textbf{48.1}  & $\epsilon$=0.1 & 28.5    & \textbf{41.0} \\ \hline
		\multirow{2}{*}{PGD}        &$\epsilon$=0.05          & 68.3        & \textbf{86.4}    &$\epsilon$=0.01           &36.4           & \textbf{70.9}    &$\epsilon$=0.01          
		& 66.0         & \textbf{88.4}     \\		&$\epsilon$=0.1          & 30.8          & \textbf{62.5}     &$\epsilon$=0.02           &12.0           & \textbf{39.2}      &$\epsilon$=0.02          &35.3          & \textbf{68.5}     \\ \hline
		Deepfool              & -        & 3.7      & \textbf{91.4} & -         & 0.4       & \textbf{84.9}  & -        & 1.1      & \textbf{84.1} \\ \hline
		\multirow{2}{*}{C\&W} & c=0.1    & 43.7     & \textbf{83.9} & c=0.01    & 41.7      & \textbf{91.7}  & c=0.01   & 38.8      & \textbf{92.3} \\
		& c=1      & 13.8      & \textbf{75.3} & c=0.1     & 16.4       & \textbf{87.9}  & c=0.1    & 9.5      & \textbf{75.7} \\ \hline
		\multirow{2}{*}{EAD} & c=0.1    & 35.9     & \textbf{68.0} & c=0.01    & 26.2      & \textbf{79.0}  & c=0.01   & 29.8      & \textbf{80.3} \\
		& c=1      & 12.8      & \textbf{48.3} & c=0.1     & 10.8       & \textbf{47.6}  & c=0.1    & 5.9      & \textbf{69.2} \\ \hline
	\end{tabular}}
	\label{tab:1}

\end{table}
\begin{table}[t]
	\centering
	\caption{Classification accuracy (\%) on adversarial examples. Models consist of VGG-16.}
	\scalebox{0.8}{
	\begin{tabular}{c|c|c|c|c|c|c|c|c|c}
		\hline
		& \multicolumn{3}{c|}{\textbf{MNIST}} & \multicolumn{3}{c|}{\textbf{CIFAR-10}} & \multicolumn{3}{c}{\textbf{SVHN}}  \\
		Attack Method               & Para.    & Baseline & RSL           & Para.     & Baseline  & RSL            & Para.    & Baseline & RSL           \\ \hline\hline
		No Attack & - & 98.2 & \textbf{99.1} & - & 92.1 & \textbf{93.7} & - & 96.5 & \textbf{97.0} \\ \hline
		\multirow{2}{*}{FGSM} & $\epsilon$=0.4  & 67.3     & \textbf{84.3} & $\epsilon$=0.1  & 45.4      & \textbf{53.3}  & $\epsilon$=0.1 & 27.8     & \textbf{40.1} \\
		& $\epsilon$=0.8  & 16.7    & \textbf{28.6} & $\epsilon$=0.2  & 22.0      & \textbf{28.1}  & $\epsilon$=0.2 & 17.4    & \textbf{26.8} \\ \hline
		\multirow{2}{*}{PGD}        &$\epsilon$=0.1          & 66.55         & \textbf{83.9}     &$\epsilon$=0.01           &54.7           & \textbf{78.8}      &$\epsilon$=0.02          &42.5          & \textbf{74.0}     \\		&$\epsilon$=0.2          & 23.57         & \textbf{49.1}     &$\epsilon$=0.02           &37.4           & \textbf{58.9}      &$\epsilon$=0.04          &18.4          & \textbf{45.0}     \\ \hline
		Deepfool              & -        & 1.7      & \textbf{82.2} & -         & 0.9       & \textbf{67.9}  & -        & 1.3      & \textbf{68.4} \\ \hline
		\multirow{2}{*}{C\&W} & c=0.1    & 58.5     & \textbf{89.5} & c=0.01    & 61.0      & \textbf{91.5}  & c=0.01   & 47.1      & \textbf{88.6} \\
		& c=1      & 15.2      & \textbf{65.8} & c=0.1     & 18.2       & \textbf{74.3}  & c=0.1    & 13.9      & \textbf{68.5} \\ \hline
		\multirow{2}{*}{EAD} & c=0.1    & 48.1     & \textbf{65.7} & c=0.01    & 51.4      & \textbf{84.3}  & c=0.01   & 18.0      & \textbf{67.4} \\
		& c=1      & 22.2      & \textbf{45.6} & c=0.1     & 28.5       & \textbf{44.9}  & c=0.1    & 3.8      & \textbf{61.2} \\ \hline
	\end{tabular}}
	\label{tab:2}

\end{table}
We evaluate the performance against well-known white-box attacks. We apply five adversarial attack methods (FGSM \cite{goodfellow2014explaining}, PGD \cite{madry2018towards}, Deepfool \cite{moosavi2016deepfool}, C\&W \cite{carlini2017towards}, and EAD \cite{chen2018ead}) on three datasets. For the adversarial attack setting, the iteration step is set to be 10 for PGD with step size of  $\epsilon/10$ where $\epsilon$  denotes a magnitude of noise. For the C\&W and EAD, we perform with constant \textit{c}.  With a selected \textit{c}, we then run 1000 iterations of gradient decent with Adam optimizer with the learning rate of 0.01.

Table 1 and 2 show the classification accuracy on each adversarial example on ResNet-50 and VGG-16, respectively. We conduct experiment with various   settings. "No Attack" denotes the normal setting when testing with normal data. In the case of baseline, we train a sub-model and attacking that sub-model. In the case of our method, we sample a model from model set by proposed random layer sampling and attack the sampled network. Then, we evaluate accuracy by mean of 10 sample output (\textit{M}=2, \textit{N}=10). As shown in the tables, our method significantly improves adversarial robustness compared to baseline. Especially, in the case of recently proposed powerful attack methods C\&W and EAD, the accuracy of the baseline is significantly dropped on three datasets. In the case of our method, although the adversarial attack is powerful, the accuracy does not drop significantly. Although the attacker knows the full structure and the weight of the sampled model, it attacks different sampled model. With the proposed random layer selection method, we can effectively defend adversarial examples with various sampled model.

\begin{table}[t]
	\centering
	\caption{Classification accuracy comparison with other defense methods on CIFAR-10.}
	\scalebox{0.9}{
	\begin{tabular}{c|c|c|c|c|c}
		\hline
		& \multicolumn{5}{c}{\textbf{CIFAR-10}}                                                      \\
		Defense            & FGSM           & PGD            & Deepfool       & C\&W           & EAD            \\ \hline\hline
		No defense         & 45.4          & 36.4          & 0.4         & 41.7          & 26.2         \\ \hline
		ADP \cite{pang2019improving}                & \textbf{70.2}              & 61.7              &  20.3             & 52.0              & 61.9              \\ \hline
		RSE \cite{liu2018towards}                & 51.8          & 45.8             & 82.1          & 91.2          & 75.4          \\ \hline
		Stochastic Dropout \cite{dhillon2018stochastic} & 58.5          & 59.0           & 56.4           & 53.4           & 38.1           \\ \hline
		Ours          & 65.1 & \textbf{70.9}          & \textbf{84.9} & \textbf{91.7} & \textbf{79.0} \\ \hline
	\end{tabular}}
	\label{tab:3}

\end{table}
\begin{table}[t]
\centering
\caption{Classification accuracy comparison on CIFAR-10 dataset when the attacker attacks multiple sampled models.}
\begin{tabular}{ccccc}
\hline
Model                      & \begin{tabular}[c]{@{}c@{}}\# of Attacked\\ Sampled Model\end{tabular} & FGSM & PGD  & C\&W \\ \hline\hline
\multirow{4}{*}{Resnet-50} & 1                                                                      & 65.1 & 70.9 & 91.7 \\  \cline{2-5}
                           & 5                                                                      & 60.5 & 65.3 & 84.5 \\ \cline{2-5}
                           & 10                                                                     & 58.7 & 60.9 & 82.3 \\ \cline{2-5}
                           & 15                                                                     & 63.1 & 60.2 & 81.7 \\ \hline
\end{tabular}
\label{tab:4}

\end{table}
\subsection{Performance Comparison with Other Defense Methods}
We compare our method with other defense methods. We use Resnet-50 network as backbone and test on CIFAR-10 dataset. Table 3 shows classification accuracy comparison with other ensemble based defense methods. We set the attack parameters ( $\epsilon$ , \textit{c}, and \textit{number of iteration}) same as section 4.2. As shown in the table, our method outperforms other ensemble methods. Compared with other defense methods, our proposed method improves ensemble diversity through group optimization strategy. Also the sampled models can hide the gradient through random layer sampling. Through the experiment, we prove that the proposed method effectively improves the robustness against to adversarial attack.
\subsection{Multiple Attack and Defense}
If the attacker knows that the model consists of more than two models, the attacker could attack multiple models. In the multiple attacker scenario, our method could still operate robustly. To verify that our method is also robust to multiple attacks, we conduct multiple attack and defense scenario. Table 4 shows the results of the proposed method when the attacker attacks sampled models. As shown in the table, there are only few accuracy drops even the attacker attacks multiple attacks. In the case of FGSM, since it is hard to generate adversarial example that could attack more than 10 sampled model, the attacks do not work properly. Also, it is impossible to attack all sampled network. Therefore, it can be interpreted that our method is also robust to multiple attacks.

\subsection{Ensemble Diversity Comparison}
One of the main contributions of our method is to guarantee the ensemble model diversity. To verify this, we measure the diversity of the proposed method by using Interrater Agreement (IA) score \cite{kuncheva2003measures, zhang2019confidence}. This score explicitly quantifies the diversity of ensemble models. The lower the IA, the more diverse the predictions of the models. In the case of Stochastic Dropout, RSE, and our method, we use 10 sample and repeat 10 times. Since the ADP method use fixed model, we implement only one time. Table 5 shows the IA scores comparison with other ensemble methods. As shown in the table, our method shows lower IA score than other methods. It means that our proposed method guarantees the diversity.

\begin{table}[t]
\centering
\caption{Interrater Agreement (IA) score comparison with other ensemble methods.}
\scalebox{0.9}{
\begin{tabular}{ccccc}
\hline
{ Dataset}  & { Stochastic Dropout} & {RSE}          & {ADP}  & {Ours}         \\ \hline \hline
{ MNIST}    & {0.53 $\pm$ 0.005}       & {0.55 $\pm$ 0.015} & {0.52} & \textbf{0.43 $\pm$ 0.021} \\ \hline
{CIFAR-10} & {0.58 $\pm$ 0.004}       & {0.60 $\pm$ 0.031} & {0.54} & \textbf{0.51 $\pm$ 0.018} \\ \hline
{SVHN}     & {0.62 $\pm$ 0.003}       & {0.66 $\pm$ 0.024} & {0.62} & \textbf{0.58 $\pm$ 0.023} \\ \hline
\end{tabular}}
\label{tab:5}

\end{table}
\begin{table}[!t]
	\centering
	\caption{Classification accuracy on adversarial examples according to the number of sample.}
	\scalebox{0.9}{
		
	\begin{tabular}{c|c|c|c|c}
		\hline
		 Method     & \# of Samples & FGSM & PGD & C\&W \\ \hline\hline
		\multirow{2}{*}{Base ensemble} 
	    & 3        & 50.2    & 53.7        &    46.7   \\
		& 5       & 51.1    & 55.2        &  47.6  \\
		\hline
		\multirow{3}{*}{Our ensemble} 
	    & 5        & 64.8    & 70.3           & 91.5    \\
		& 10       & \textbf{65.2}    & 70.8           & \textbf{91.7}    \\
		& 15       & 65.1    & \textbf{70.9}           & \textbf{91.7}    \\
		\hline
	\end{tabular}}
	\label{tab:6}

\end{table}
\subsection{Effect of Number of Sample for Adversarial Robustness}
We analyze the effect of the number of samples for adversarial robustness. Table 6 shows the classification accuracy according to the number of samples. In the case of the Base ensemble, we construct 3 and 5 ResNet-50 sub-models and train individually. Then, we select one sub-model and generate adversarial examples. As shown in the table, as the number of samples increase, the adversarial robustness is also increased. However, to improve the robustness, a large number of parameters are required. On the contrary, in the proposed method, with only two sub-models, it is possible to generate various sampled models with random layer sampling. As a result, our method can hide the weight of the attacked sampled-model effectively. Therefore, our method effectively defense the adversarial examples with a small number of sub-models.

\section{Conclusion}
This paper presents an ensemble model training framework with a random layer sampling method for adversarial robustness. In the proposed method, we design a model set consist of multiple sub-models and construct sampled models with the random layer sampling method. The sampled models are trained by a group optimization strategy to guarantee diversity. After the training, our method predicts various sample outputs by recombination of layers. Therefore, our method effectively hides the full gradient of the models and improves the adversarial robustness. Comprehensive and comparative experiments show that the proposed method could defend adversarial attacks effectively.

\section*{Acknowledgements}
This work was conducted by Center for Applied Research in Artificial Intelligence (CARAI) grant funded by DAPA and ADD (UD190031RD).
\bibliography{egbib}
\end{document}